\documentclass[runningheads]{llncs}
\usepackage{graphicx}
\usepackage{booktabs}
\usepackage{amsfonts}
\usepackage{subfigure}
\usepackage{url}
\usepackage[linesnumbered,ruled,vlined]{algorithm2e}
\begin{document}
\title{Enabling Data Diversity: \\ Efficient Automatic Augmentation via  \\ Regularized Adversarial Training}



\titlerunning{Efficient Automatic Augmentation via Regularized Adversarial Training}
%

\author{Yunhe Gao\inst{1} \and
Zhiqiang Tang\inst{1} \and
 Mu Zhou\inst{1,2}\and
Dimitris Metaxas \inst{1}
}
\authorrunning{Y. Gao et al.}
%
\institute{Department of Computer Science, Rutgers University \and
SenseBrain and Shanghai AI Laboratory and Centre for Perceptual and Interactive Intelligence \\
}
\maketitle              
\begin{abstract}

Data augmentation has proved extremely useful by increasing training data variance to alleviate overfitting and improve deep neural networks' generalization performance. In medical image analysis, a well-designed augmentation policy usually requires much expert knowledge and is difficult to generalize to multiple tasks due to the vast discrepancies among pixel intensities, image appearances, and object shapes in different medical tasks. To automate medical data augmentation, we propose a regularized adversarial training framework via two min-max objectives and three differentiable augmentation models covering affine transformation, deformation, and appearance changes.  Our method is more automatic and efficient than previous automatic augmentation methods, which still rely on pre-defined operations with human-specified ranges and costly bi-level optimization. Extensive experiments demonstrated that our approach, with less training overhead, achieves superior performance over state-of-the-art auto-augmentation methods on both tasks of 2D skin cancer classification and 3D organs-at-risk segmentation.


\end{abstract}
\section{Introduction}

\let\thefootnote\relax\footnotetext[0]{\noindent Corresponding to \{zhiqiang.tang,dnm\}@cs.rutgers.edu and muzhou1@gmail.com. This research was supported in part by NSF: IIS 1703883, NSF IUCRC CNS-1747778 and funding from SenseBrain, CCF-1733843, IIS-1763523, IIS-1849238, MURI- Z8424104 -440149 and NIH: 1R01HL127661-01 and R01HL127661-05. and in part by Centre for Perceptual and Interactive Intellgience (CPII) Limited, Hong Kong SAR.}

Data augmentation is a crucial technique for deep neural network training and model optimization \cite{cubuk2020randaugment,krizhevsky2012imagenet,lecun1998gradient,lim2019fast}. The basic idea is to improve data amount and diversity through a set of transformations, such as image rotation, scale, deformation, and random noise \cite{lecun1998gradient}. The increased amount of training samples are valuable for the performance of computer-assisted medical systems and tasks such as skin cancer diagnosis \cite{brinker2018skin} and organs-at-risk segmentation \cite{gao2019focusnet}. 


However, there are wide variations in image intensity, appearance, and object shape for different medical tasks. A hand-crafted augmentation policy requires expert knowledge and is unlikely to generalize across multiple tasks. Furthermore, an augmentation policy usually comes with random sampling, which is not adaptive to the target model's state during training. The isolation from model training impairs the effectiveness of data augmentation. Therefore, medical image analysis faces a vital problem: {\it How to effectively augment medical data with more diversity and less human effort}?

The recent works on automatic data augmentation (i.e., AutoAugment) \cite{cubuk2018autoaugment,yang2019searching} are at the forefront to address the above challenge. Nevertheless, the formulation of AutoAugment has restricted its scope of usability in terms of automation and efficiency. First, it still relies on a set of human-specified transformation functions. Designing useful search space, i.e., defining the transformations with proper boundaries, may require domain knowledge on new tasks, leading to reduced automation and increased user burdens. Second, it uses validation data to evaluate data augmentation performance, resulting in an inefficient bi-level optimization problem, despite using proxy techniques (e.g., smaller models) \cite{cubuk2018autoaugment}.  Moreover, the quantized parameters (probability and magnitude) in AutoAugment are not differentiable, causing additional optimization difficulty.
 

To address the automation and efficiency issues, we propose a data augmentation framework via regularized adversarial training. 
To improve automation, we establish a new search space consisting of three components: global affine transformation, local deformation, and appearance perturbation, each of which is parameterized by a conditional generative network. Towards efficient training, we propose two min-max objectives, where the augmentation generators compete with both the discriminator and target learner to generate realistic (for discriminator) and challenging (for target learner) augmentations. The generators, discriminator, and target learner are optimized jointly in an end-to-end manner without using any validation data. The intuition is that automatically controlled adversarial training can reduce the target learner's overfitting and enhance its generalization power. Overall, our major contributions are:
\begin{itemize}
    \item We propose a new search space parameterized by three differentiable conditional generative networks that generate global and local geometric transformations as well as appearance perturbation with little domain knowledge.
    \item To boost training efficiency, we avoid using validation data and switch the bi-level optimization objective in AutoAugment to two min-max games \cite{goodfellow2014generative,goodfellow2014explaining}, which are carefully formulated within a triplet: an augmentation generator, a discriminator, and a target network.
    
    \item Our data augmentation framework is automatic, generalizable, and computationally efficient for a variety of data-driven tasks. Extensive experiments demonstrated its superior performance and efficiency on both 2D skin cancer classification and 3D organ-at-risk segmentation tasks.
\end{itemize}

\let\thefootnote\relax\footnotetext[0]{\noindent Project page: \url{https://github.com/yhygao/Efficient_Data_Augmentation}}

\section{Related Work}

Our analysis is conceptually related to previous data augmentation approaches,  techniques to automate data augmentation, and adversarial training on pixel perturbation and image transformation. 

\textbf{Data augmentation} applies transformations to the training samples to facilitate model training and reducing overfitting in machine learning. Previously, there were limited augmentation operations available such as random crop, scaling, flipping and rotation prepared for network training \cite{lecun1998gradient}. In the medical image field, random elastic local deformation was used for simulating anatomical variations \cite{isensee2018nnu}. Random perturbation on color or intensity values was also used to diversify the training data \cite{krizhevsky2012imagenet}. However, all of these methods are based on human-designed heuristics requiring substantial domain knowledge that are unable to extend well in unseen datasets. 

\textbf{Automatic data augmentation} is emerging to search the optimal composition of transformations to improve machine learning model performance. AutoAugment \cite{cubuk2018autoaugment} explores policies and validates their effectiveness by repeatedly training models from scratch. Due to the large search space and bi-level optimization using reinforcement learning, AutoAugment suffers  from extremely high  computational  demand. Subsequent  works  \cite{lim2019fast,cubuk2020randaugment} aim at reducing the computational complexity of the searching. In the medical image field, Yang et al. \cite{yang2019searching} proposed an automated searching approach for both data augmentation and other training strategies for 3D medical image segmentation.  \cite{xu2020automatic} propose a differentiable way to update augmentation parameters by means of stochastic relaxation and Monte-Carlo method. However, they all followed the formulation of AutoAugment and used pre-defined transformations (e.g. rotation or contrast enhancement) and search for the optimal magnitude. These randomly sampled transformations can be insufficient for capturing subtle variations in medical images and have not fully exploited the capabilities of augmentation space.

To automatically explore the transformation, researchers used spatial and appearance transform model \cite{zhao2019data} to learn the registration field and additive mask between two MR images. During model training, they considered the two models to augment training samples by matching external unlabeled data. Also, a deformation model and an intensity model \cite{chaitanya2019semi} were considered to generate realistic augmentation images. However, these methods are more complicated to use as they require multi-stage training.

\textbf{Adversarial training} is designed to train networks using samples with a small perturbation in the adversarial direction to improve the model's robustness to adversarial attack \cite{goodfellow2014explaining,madry2017towards}. Most existing works choose to add small pixel-wise perturbation to the input by adding gradient-based adversarial noise \cite{paschali2018generalizability,tramer2019adversarial}. Some works use GAN to approximate the distribution of adversarial examples and generate adversarial perturbations \cite{xiao2018generating}.  Recent findings suggested that spatial transformations, such as affine transformations \cite{engstrom2019exploring}, elastic deformations \cite{alaifari2018adef}, can help produce adversarial examples. Our study draws inspiration from adversarial training, but we focus on increasing the generalization capability and improving the performance of clean data instead of adversarial samples.

\section{Method}

\begin{figure}[t]
\setlength{\abovecaptionskip}{0pt} 
\setlength{\belowcaptionskip}{-10pt}

\begin{minipage}[]{.4\textwidth}
\centering
\includegraphics[scale=0.45]{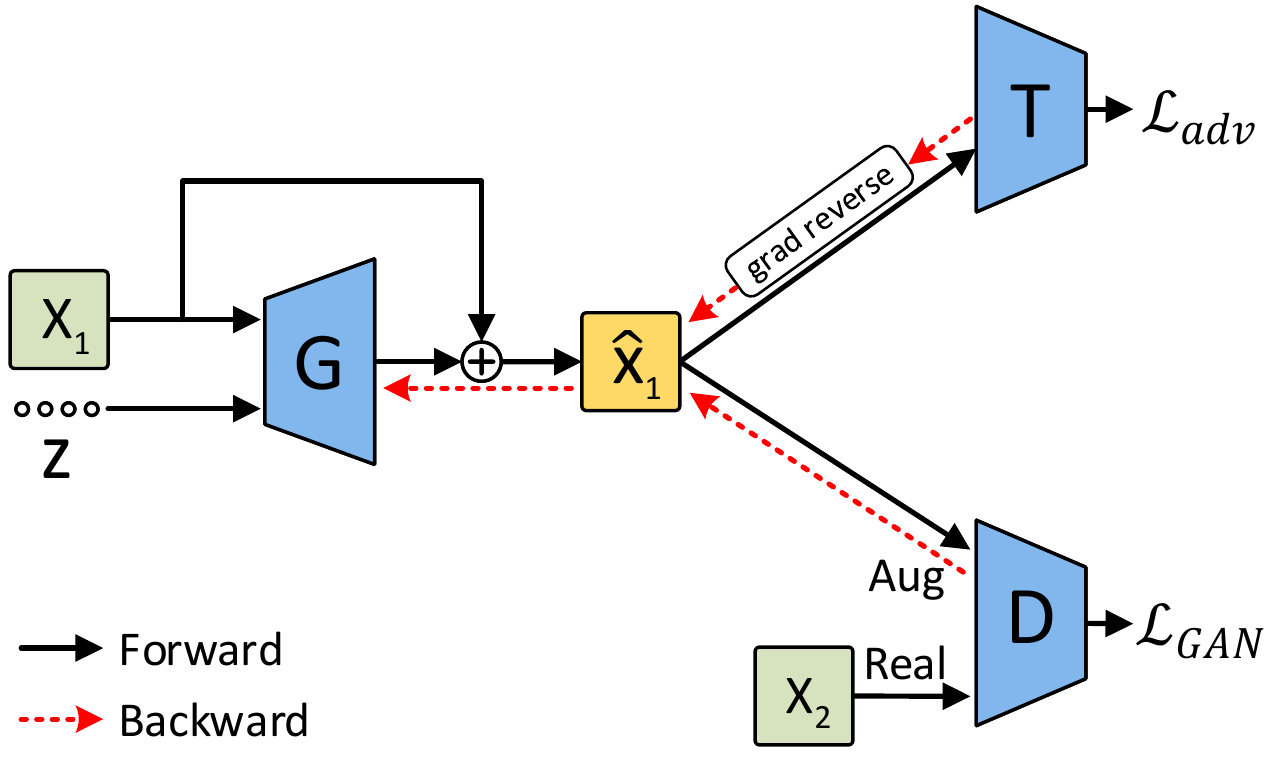}
\end{minipage}\hfill
\begin{minipage}[]{.53\textwidth}
\begin{scriptsize}
\begin{algorithm}[H]
\SetAlFnt{\scriptsize}
\SetAlCapFnt{\scriptsize}

\hsize=\textwidth
\SetAlgoLined
\KwIn{Initial target model $T$, augmentation generator $G$, discriminator $D$}
\While{not converged}{


Sample batches $\mathbf{x}_{1},\mathbf{x}_2$ and noise vector $z$;

Generate augmented images: $\mathbf{\hat{x}}_1=G(z,\mathbf{x}_1)\oplus \mathbf{x}_1$;

Compute grad w.r.t. $T$ and $G$ with $\mathbf{\hat{x}}_1$ to minimize Eq. \ref{eq:overall}. The gradient from $T$ is reversed to optimize $G$ and $T$ jointly;

Compute grad w.r.t. $D$ with $\mathbf{x}_2$ and $\mathbf{\hat{x}}_1$ to maximize Eq. \ref{eq:adv2};

Update $T, G$ and $D$;

}
\KwOut{Optimized target model $T^*$}
\caption{\scriptsize Optimization scheme} \label{alg:algorithm}
\end{algorithm}
\end{scriptsize}
\end{minipage}

\caption{The regularized adversarial data augmentation framework ({\bf Left}) and its training algorithm ({\bf Right}). Augmentation generator $G$
plays two min-max games with target model $T$ and discriminator $D$ simultaneously to generate challenging and realistic augmented images, hence reduced overfitting and improved generalization. }\label{fig:framework}
\end{figure}


\subsection{Preliminaries}

In the vanilla supervised training setting with data augmentation, the objective function is

\begin{equation}
     \mathop{\arg\min}_{T}\ \mathbb{E}_{x,y\in \mathrm{\Omega}} \ \mathcal{L}( \phi(x), y; T)\label{eq:target}
\end{equation}
where $\mathcal{L}(\cdot,\cdot;\cdot)$ is the loss function, $T$ and $\phi$ are the target and augmentation models; $x$ is a sample from the training dataset $\mathrm{\Omega}$; and $y$ is its corresponding ground-truth label. For example,  in the skin lesion diagnosis task, $x,y$ are dermoscopic images and the corresponding lesion categories; and $T$ is a classifier with cross-entropy loss $\mathcal{L}$. Note that for some tasks like segmentation, a spatial transformation should be applied to $x$ and $y$ simultaneously.



In hand-crafted data augmentation, $\phi$ includes a set of pre-defined transformation operations with manually designed boundaries and sampling probabilities. AutoAugment \cite{cubuk2018autoaugment} moves one step further by searching for the transformation magnitudes, probabilities, and combinations. However, it still depends on the pre-defined operations and human-specified boundaries. In our approach, $\phi$ are learnable generative neural networks, with broad, differentiable, and continuous search space.


\subsection{Regularized Adversarial Data Augmentation}

The key idea of our approach is to use adversarial training to automate data augmentation. In Fig. \ref{fig:framework}, we propose the regularized adversarial training framework including three networks: an augmentation generator $G$, a target network $T$, and a discriminator $D$ and two min-max objectives. In the spirit of adversarial training, augmentation generator $G$ competes with target network $T$:
\begin{equation}
     \mathcal{L}_{adv} = \mathop{\min}_{T}\ \mathop{\max}_{G}\  \mathbb{E}_{x,y\in \mathrm{\Omega}} \ \left[\mathcal{L}(G(z, x)\oplus x, y; T)\right], \label{eq:adv1}
\end{equation}
where $G$ is a conditional generative network that generates augmentation transformations based on the original image $x$ and the Gaussian noise vector $z$. The wrapping operation $\oplus$ varies for different augmentation models, whose details will be introduced in Section \ref{sec:aug-models}. The above objective Eq. \ref{eq:adv1} allows the augmentation model to search along the adversarial direction, such that it generates challenging augmentations to let the target model learn robust features, thus alleviate the potential overfitting issue. 


Without proper constraint, adversarial training probably leads to excessive augmentations that hurt the model generalization on clean data \cite{ilyas2019adversarial,tsipras2018robustness}. To avoid this, we offer two regularization terms that work in the augmentation and image spaces, respectively. First, we regularize the augmentation magnitude by:
\begin{equation}
    \mathcal{L}_{reg} = \mathop{\min}_{G}\ \mathbb{E}_{x\in \mathrm{\Omega}} \mathcal{R}(G(z, x)), \label{eq:reg}
\end{equation}
 which aims to penalize large augmentations since they probably move the original data out of distribution. We will elaborate $\mathcal{L}_{reg}$ in Section \ref{sec:aug-models} as it is related to concrete augmentation models. To further improve the augmentation quality, we use discriminator $D$ to encourage augmented data to look realistic. Thus, we propose the second adversarial objective:
\begin{equation}
    \mathcal{L}_{GAN} = \mathop{\min}_{G}\ \mathop{\max}_{D}\ \mathbb{E}_{x\in \mathrm{\Omega}} \left[\log D(x)\right] + \mathbb{E}_{x\in \mathrm{\Omega}} \left[\log (1-D(G(z, x)\oplus x))\right] \label{eq:adv2}
\end{equation}
Combining Eq. \ref{eq:adv1}, \ref{eq:reg}, and \ref{eq:adv2} gives the overall objective:
\begin{equation}
    \mathcal{L}_{overall} = \mathcal{L}_{adv} + \lambda \mathcal{L}_{GAN}+\gamma \mathcal{L}_{reg}, \label{eq:overall}
\end{equation}
where $\lambda$ and $\gamma$ represent loss weights.

Our regularized adversarial data augmentation plays two min-max games within a triplet of $G$, $D$, and $T$. Augmentation generator $G$, considering both input data and target network $T$'s training state, aims to generate challenging augmentations for target learner $T$. Simultaneously, it still needs to ensure that the augmented data are in-distribution and achievable by regularizing the augmentation magnitude and forcing the augmented data to be realistic by discriminator $D$. Using the discriminator is beneficial because small transformations may still make augmented data look fake. By competing with $T$ and $D$, $G$ can automatically learn data augmentation and thus require little human effort. 

The optimization of the triplet is efficient in an end-to-end manner. In each iteration of Algorithm \ref{alg:algorithm}, we sample two mini-batches $x_1$ and $x_2$ from the training data. $x_1$ and the sampled Gaussian noises $z$ will go through augmentation generator $G$ to become augmented $\hat{x}_1$, which is then fed into target learner $T$ and discriminator $D$. Note that we reverse the gradient from $T$ before it flows into $G$, such that we can compute the gradient with respect to $T$ and $G$ at the same time to minimize Eq. \ref{eq:overall} through one backward call. Then $\hat{x}_1$ and $x_2$ are used to compute the gradient of $D$ to maximize Eq. \ref{eq:adv2}. Thus we can update the triplet with only one pair of forward and backward flow. The steps of training the triplet are given in Algorithm \ref{alg:algorithm}.






\subsection{Data Augmentation Space and Models}\label{sec:aug-models}

\begin{figure}[t]
\setlength{\abovecaptionskip}{0pt} 
\setlength{\belowcaptionskip}{-10pt}
\centering
\includegraphics[width=0.85\linewidth]{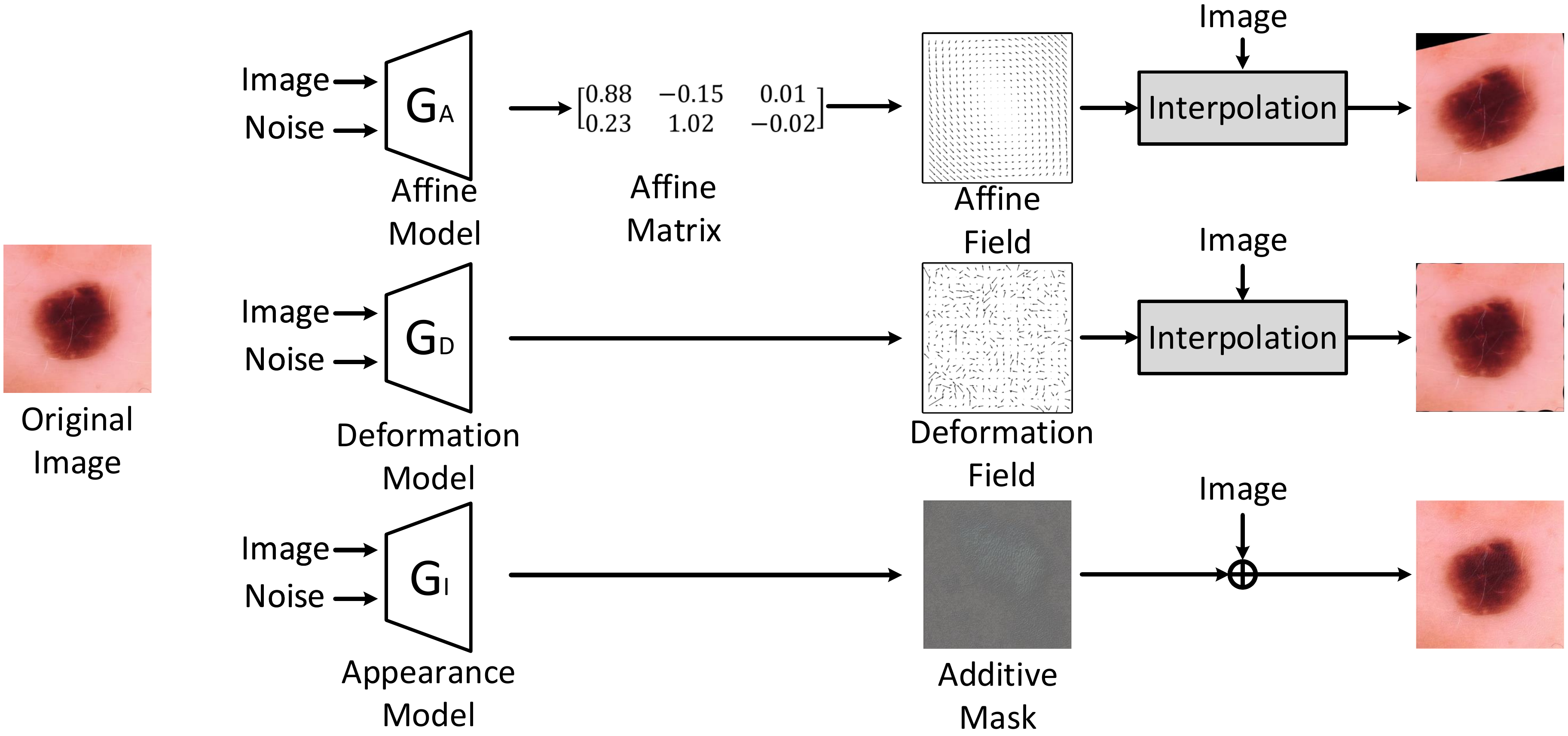}

\caption{Illustration of the three augmentation models which perform affine transformation ({\bf Top}), local deformation ({\bf Middle}), and appearance transformation ({\bf Bottom}). Each model is conditioned on an input image and a Gaussian noise vector and generates an affine matrix, a deformation field, or additive appearance mask. }\label{fig:aug_net}
\end{figure}

We characterize the medical image variations by three types of transformations: affine transformation, local deformation, and appearance perturbations.  These transformations commonly exist for a wide range of medical tasks \cite{brinker2018skin,isensee2018nnu,yang2019searching}. Instead of relying on expert knowledge to specify the transformation ranges, we use three convolutional neural networks ($G_A,\ G_D$, and $G_I$) to learn to generate transformations adaptively. The learning is fully automatic when plugging them into the above regularized adversarial data augmentation framework. Each transformation generator, conditioned on an input image $x$ and a Gaussian noise vector $z$, produces an affine field, deformation field, or additive appearance mask. Figure \ref{fig:aug_net} illustrates the three transformations and we introduce details below.

We use affine transformation to approximate the global geometric transformation of an image. In the case of 2D images, our affine model $G_A$ can generate a $2\times3$ affine matrix $A$, representing an affine transformation. To make the affine transformation differentiable, we turn to the spatial transformer module \cite{jaderberg2015spatial} that first converts affine matrix $A$ to a flow field denoted as $G_A(z, x)$, and then conducts differentiable image sampling $\oplus$ to generate a transformed image $\hat{x}=G_A(z, x) \oplus x$. We regularize the magnitude of flow field $G_A(z, x)$ as follows:
\begin{equation}
    \mathcal{L}_{reg}^A=||G_A(z,x)-f_{id}||_2,
\end{equation}
where the $f_{id}$ is the identity flow field indicating no transformation.



Complementary to global affine transformation, deformation can characterize local geometric transformations. Deformation model $G_D$ directly generates a residual flow field $G_D(z, x)$ including pixel-wise displacements. Then we add identity flow field $f_{id}$ on $G_D(z, x)$ to obtain the deformation field and then use the differentiable sampling $\oplus$ to obtain the deformed image $\hat{x}=(G_D(z, x)+f_{id}) \oplus x$. Here we encourage the deformation smoothness by the following regularization:

\begin{equation}
    \mathcal{L}_{reg}^D = ||\nabla G_D(z, x)||_2,
\end{equation}
where $\nabla$ is the spatial gradient operator to calculate the differences between neighbor elements in the deformation field $G_D(z, x)$.



Orthogonal to geometric transformations, appearance transformation perturbs the pixel intensity. For simplicity, we transform image appearance by pixel-wise addition. Appearance transformation model $G_I$ outputs the appearance mask $G_I(z, x)$ to generate the perturbed image $\hat{x}=G_I(z, x)+x$. We use the following regularization loss:

\begin{equation}
    \mathcal{L}_{reg}^I = ||G_I(z,x)||_2,
\end{equation}
which can constrain the perturbation magnitude of $G_I(z, x)$.




To combine the three augmentation models, we draw inspirations from AdvProp \cite{xie2020adversarial} to use three groups of auxiliary batch normalizations (BNs) in the target network to capture three augmentation distributions separately, and a main group of BNs for the clean data. In mini-batch training, we split a batch into four parts, three of which are processed by the three augmentation models to generate augmented data. They are fed into the target learner $T$ using their corresponding BNs, and then we update the triplet, as illustrated in Algorithm \ref{alg:algorithm}. When testing the trained target network, we only use the main BNs.

\vspace{-10pt}

\section{Experiments}

\subsection{Experiments Setup}

\textbf{Datasets.}
We performed experiments on a 2D image classification task and a 3D image segmentation task. 
\begin{itemize}
  \item[$\bullet$] Skin lesion diagnosis. We used the public ISIC 2018 challenge dataset \cite{codella2019skin} to make automated predictions of 7 categories of skin lesion within 2D dermoscopic images. The training dataset consists of 10,015 skin lesion images with a fix size of $450\times 600$ pixels.
  
  \item[$\bullet$] Organ-at-risk (OAR) segmentation. We used OARs dataset in MICCAI 2015 challenge to segment 9 OARs in head-and-neck region from 3D CT images. It consists of 38 CT scans for training and 10 scans for testing.
\end{itemize}

\noindent
\textbf{Implementation details.}  The three augmentation generators share similar architectures, which take an image and a 128-dimension Gaussian noise vector as inputs, but with different outputs. They are similar to the generator in DCGAN \cite{radford2015unsupervised} but consist of two branches to process the inputs. The Gaussian noise vector is reshaped to a feature map and processed by six convolutional layers, interleaved with upsampling layers and is finally upsampled to the same size as the image. The image branch also outputs a feature map with the same size through 4 convolutional layers. The two feature maps are concatenated together and then is further processed with 4 convolutional layers to the output layer. $G_A$ first global average pool the feature map and use a fully-connected layer to output the affine matrix. The output layer of $G_D$ and $G_I$ are convolutional layers followed the tanh activation function. All convolutional layers are followed by BN and ReLU except for the output layer. The three generator shares a discriminator, which consists of a series of convolutional layers, BN, and LeakyReLU with slope 0.2, and outputs the final probability through a Sigmoid activation function. For 3D tasks, all networks are modified to theirs 3D counterpart. We set $\lambda=1$ and $\gamma=0.1$ in Eq. \ref{eq:overall} in the experiments.

\noindent
\textbf{Comparison methods.}
For 2D classification, we considered multiple classification backbones, including Efficientnet B0-B2 \cite{tan2019efficientnet}, ResNet34-50 and DenseNet121, to compare the effectiveness of data augmentation methods. First, we trained the classification network without data augmentation, named "no-aug" as the lower bound. We applied three strong auto-augmentation methods as comparison methods. RandAug \cite{cubuk2020randaugment} applies a simplified search space for efficient data augmentation containing only two hyper-parameters. Fast-AutoAug \cite{lim2019fast}
searches effective augmentation policies based on density matching without back-propagation for network training. \textit{cGAN} \cite{chaitanya2019semi} is a task-driven data augmentation approach. All methods train the target network using the same setting for 70 epochs.

For 3D segmentation, we used the 3D ResUNet as the backbone network. The baseline method was trained without any data augmentation except for random cropping. We compareed our method with a commonly used manually designed augmentation policies used in the nnUNet framework \cite{isensee2018nnu}, which is wrapped as a Python image library for network training\footnote{https://github.com/MIC-DKFZ/batchgenerators}. It consists of scaling, rotation, elastic deformation, gamma transformation, and additive noise. We used all the transformation with the default parameters defined in the library (named as \textit{nnUNet Full}). We also tried to remove the gamma transformation and train the network (named as \textit{nnUNet Reduce}). The approach in \cite{chaitanya2019semi} was also extended to 3D as a comparison method (named as \textit{cGAN}).

\vspace{-8pt}
\subsection{Skin Lesion Diagnosis Result}
\vspace{-5pt}

 We evaluated each method's performance using the average recall score (also known as balanced multi-class accuracy (BMCA)) and reported the accuracy of 5-fold cross-validation in the following sections.

\begin{figure}[t]
\setlength{\abovecaptionskip}{5pt} 
\setlength{\belowcaptionskip}{-10pt}
\begin{minipage}[]{.4\linewidth}
\includegraphics[width=6.5cm]{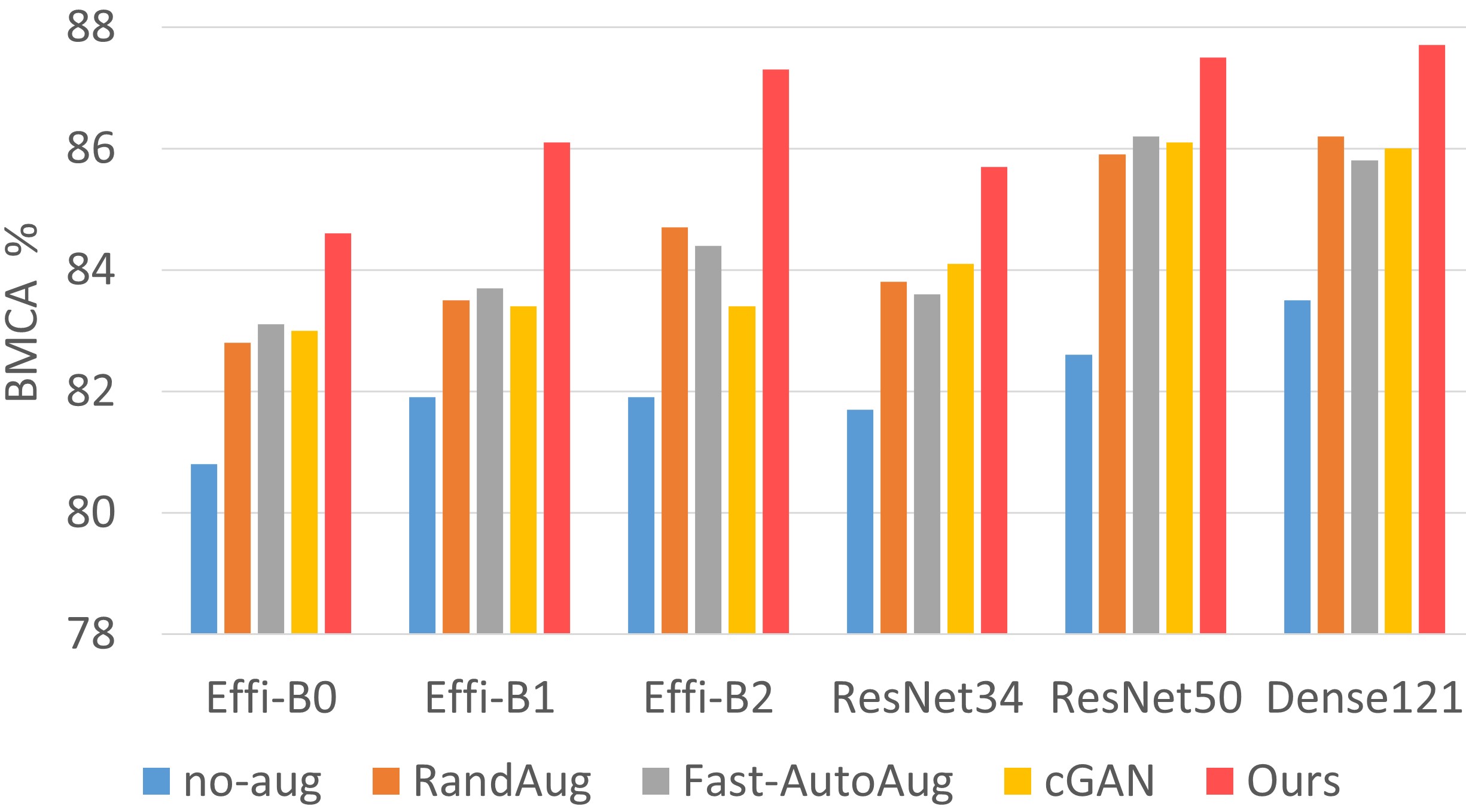}
\end{minipage}
\hfill
\begin{minipage}[]{.45\linewidth}
\begin{center}
 \scriptsize
\setlength\tabcolsep{2pt}

\begin{tabular}{l|c|c|c}
\toprule
        & W/o image & W/o & W/ \\
        & condition & discriminator & both\\
\hline
Affine  &0.837 &0.833  &0.844\\ \hline
Deform.  &0.830 &0.824  &0.832\\ \hline
Appear.  &0.827 &0.819  &0.831\\ \hline
Comb. &0.864 &0.859  &0.873\\

\bottomrule
\end{tabular}
\end{center}
\end{minipage}
\caption{Comparisons of different data augmentation methods with multiple target classifiers in the skin lesion task ({\bf Left}) and ablation studies on image condition and discriminator usage in our method using the Efficientnet-B2 classifier ({\bf Right}). Our data augmentation can notably outperform other methods in increasing classification accuracy. Additionally, the image condition and discriminator usage are beneficial because they can help produce customized transformations and realistic augmented images. }\label{isic_result}
\end{figure}


\noindent
\textbf{Results.}
Figure \ref{isic_result} shows the results of different classification backbones with data augmentation methods. Our method achieved leading performance across all backbones, especially in the Efficientnet series. Also, all automatic data augmentation methods had a large performance gain compared with no augmentation. Among Efficientnet backbones, the model complexity goes up from B0 to B2, but the increase of complexity did not always improve performance. Without data augmentation, Efficientnet-B2 resulted in similar performance with Efficientnet-B1. We found that data augmentation can further release the large models' potential, and our model obtained more improvement to Efficientnet-B2 compared to the results of RandAug and Fast AutoAug.

\begin{figure}[t]
\setlength{\abovecaptionskip}{0pt} 
\setlength{\belowcaptionskip}{-10pt}
\centering
\includegraphics[width=0.9\linewidth]{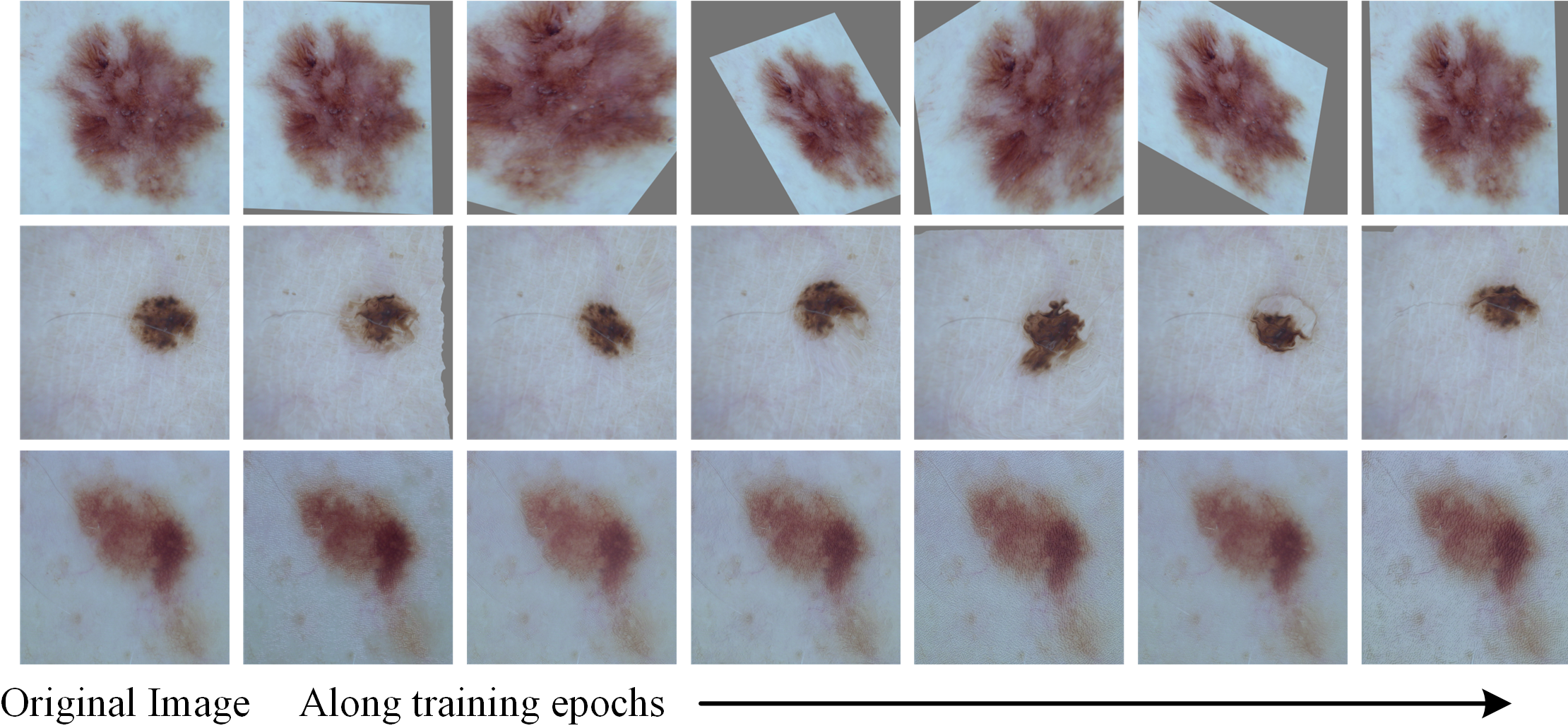}

\caption{Visualization of augmented skin lesion images along six sampled training epochs. Our affine transformation ({\bf Top}), deformation ({\bf Middle}), and appearance perturbation ({\bf Bottom}) models can generate diverse augmentations. For better visualization quality, please see the GIF in the project page.}\label{fig:isic_result_vis}
\end{figure}


Figure \ref{fig:isic_result_vis} highlighted representative augmented images from our augmentation models. Strikingly, we observed that the deformation and appearance transformations are highly centered around the lesion region, which is valuable for boosting performance of deep-learning classifiers. As opposed to traditional random deformation, the strength of our models is that they were adversarially trained with the target model and conditioned upon the input image. Therefore, our approach can generate semantically hard examples for the target model, enabling rich information extraction for the region-of-interest.

\noindent
\textbf{Ablation study.} Figure \ref{isic_result} (right) presented the ablation study of each component of our proposed method on Efficientnet-B2 backbone. Among the three proposed augmentation models, the affine model performed the best when they were applied separately. After combining them, the performance was further boosted. We also conducted experiments on the input to see if the generator conditioned on the image is useful, see the results in \textit{'W/o image condition'}. The generator's architecture was the same as the previous one, except that the image branch was removed. The results showed that conditioning on image can let the generator samples more meaningful image transformations. \textit{'W/o discriminator'} presented the results without the discriminator.  Although regularization loss can constrain the magnitude of augmentation, the discriminator make the generated data more realistic from the image space and therefore have better performance.

\noindent
\textbf{Efficiency analysis.} Table \ref{tab:gpu_hour} exhibited the GPU hours that each method consumed to search augmentation policy for the ResNet-50 backbone. The efficiency of our approach was exemplified by the joint optimization with the target model, where no retraining is needed. Our single model only took approximately 1.57x time than the original 'no-aug' training, while using the combination of three models only takes up to 2.07x time. By comparison, Fast-AutoAug searched the augmentation policies (5.5 hours) and continued to retrain the target network (3 hours). RandAug reduced the search space to two parameters, but required to train the target network 12 times. cGAN \cite{chaitanya2019semi} had two transformation models via pre-training (about 4.5 hours respectively) and then were fixed to retrain the target network using both original and augmentation data (about 9.7 hours).


\begin{table}[t]
\centering
 \footnotesize
\setlength\tabcolsep{4pt}
\caption{Training hours when using different augmentation methods with ResNet-50 in the skin lesions diagnosis. We use the NVIDIA RTX8000 GPUs and report the total training time, including updating both ResNet-50 and augmentation networks/policies.}\label{tab:gpu_hour}
\begin{tabular}{l|c|c|c|c|c|c}
\toprule
Method      &no-aug & Fast AA  &RandAug &cGAN \cite{chaitanya2019semi}   &Ours Single  &Ours Combined            \\
\hline
GPU hours      & 3     & 8.5           &36     &18.7      & 4.7       & 6.2   \\
\bottomrule
\end{tabular}
\end{table}

\vspace{-10pt}
\subsection{Organ-at-risk Segmentation Result}
\vspace{-3pt}

We extended to conduct experiments on a different OAR segmentation task using 3D CT images to reaffirm the high-level performance of our approach. We used the average Dice score over all organs as the evaluation metric.



\begin{table}[t]
\centering
 \small
\setlength\tabcolsep{4pt}

\caption{Comparison to previous data augmentation methods on the OAR segmentation. Our three individual models obtain comparable performance as the strong baseline nnUNet and more advanced cGAN. 
Combining them brings the highest Dice score. }\label{oar_result}
\begin{tabular}{l|c|cc|c|cccc}
\toprule
Method      & 3D UNet    &nnUNet &nnUNet &cGAN &\multicolumn{4}{c}{Ours }   \\
            &  Baseline &Full   &Reduce  &\cite{chaitanya2019semi} &Affine&Deform&Appear&Comb              \\
\hline
AVG Dice    & 0.784 & 0.786 &0.798 &0.803 & 0.809 & 0.808  & 0.792 &0.814 \\
\bottomrule
\end{tabular}


\end{table}

\begin{figure}[t]
\centering
\setlength{\abovecaptionskip}{0pt} 
\setlength{\belowcaptionskip}{-10pt}

\includegraphics[width=0.9\linewidth]{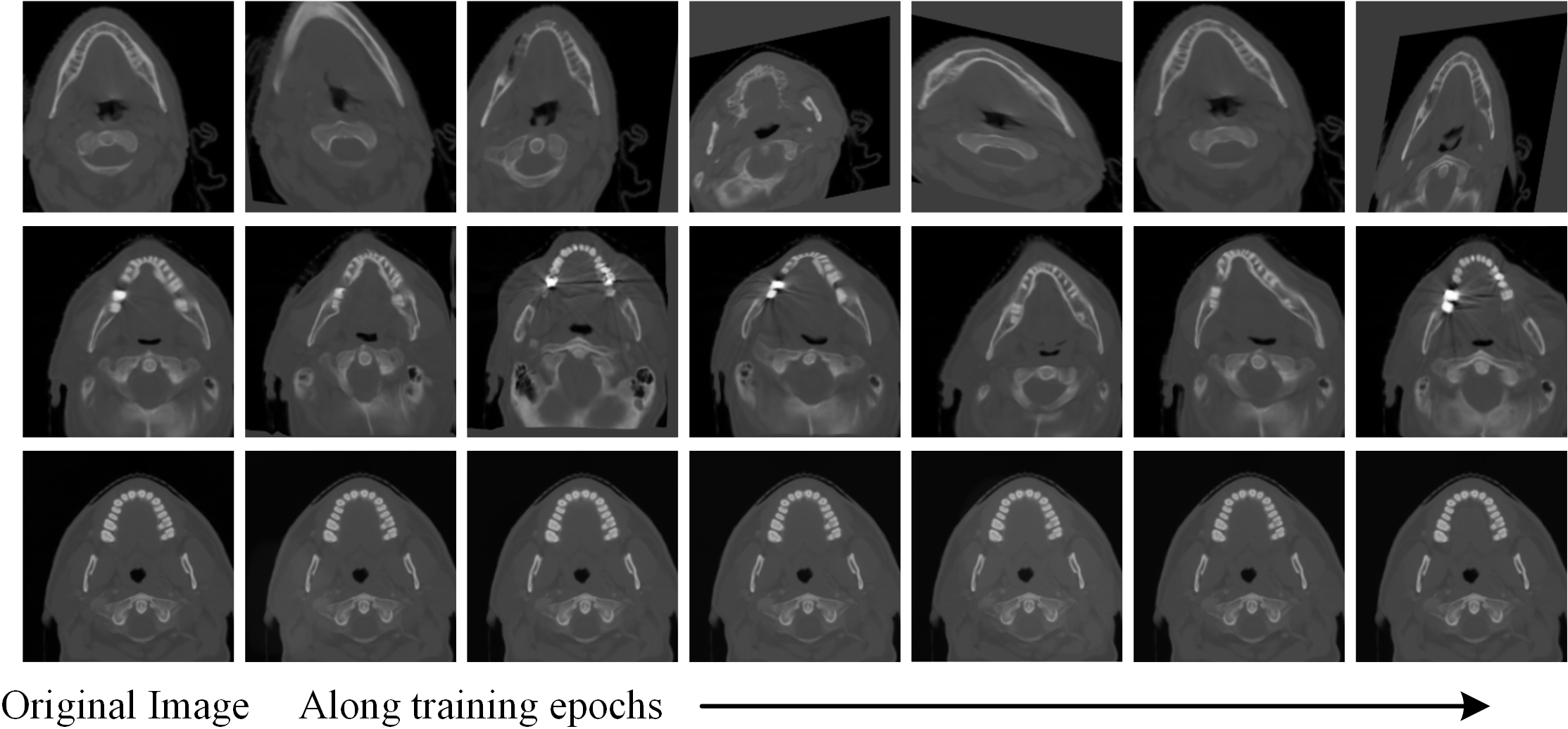}

\caption{Visualization of augmented images of the affine model (\textbf{Top}), deformation model (\textbf{Middle}) and appearance model (\textbf{Bottom}) in OAR segmentation. The augmented images present diverse variations and also preserve details of organ anatomy.}\label{fig:oar_result_vis}
\end{figure}

\noindent
\textbf{Results.} Table \ref{oar_result} showed that combination of our transformation models outperformed other competing methods. Even our single affine transformation model achieved higher performance than the cGAN \cite{chaitanya2019semi} which used pre-training to facilitate the augmentation. Notably, all of our single models demonstrated competitive performance comparing to the manually-designed \textit{nnUNet} methods. Such advance can be probably attributed to the capability of adversarial training allowing us to augment diverse training samples. Furthermore, the higher result of \textit{nnUNet Reduce} over \textit{nnUNet Full} reflected the inherent complexity of medical image tasks, suggesting that one good setting of data augmentation parameters in a task (i.e., \textit{nnUNet Full}) is unlikely to generalize well in different tasks. Fig. \ref{fig:oar_result_vis} presented the diverse variations of the intermediate slices from our augmentation models.

\vspace{-10pt}
\section{Conclusion}
\vspace{-2pt}

In this paper, we have proposed an efficient automatic data augmentation framework via regularized adversarial training. Three augmentation models were designed to learn the comprehensive distribution of augmentation for improving medical image analysis. The regularized adversarial training is crucial for boosting neural network performance and the joint optimization brings computational efficiency. We showed that the augmentation model can generate meaningful transformations that produce hard yet realistic augmented images. Extensive experiments demonstrated that our method outperformed state-of-the-art augmentation methods in both performance and efficiency. This approach could aid in many image-based clinical applications, especially when limited annotated examples are available. 

\vspace{-10pt}

%
%
%
%
\bibliographystyle{splncs04}
\bibliography{egbib}

\end{document}